\documentclass[final]{l4dc2026}

\newenvironment{myitem}{\begin{list}{$\bullet$}
{\setlength{\itemsep}{0pt}
\setlength{\topsep}{-5pt}
\setlength{\leftmargin}{12pt}
\setlength{\parsep}{0pt}
\setlength{\itemsep}{0pt}
\setlength{\partopsep}{0pt}}}%
{\end{list}}

\title[CableRobotGraphSim]{CableRobotGraphSim: A Graph Neural Network for\\ Modeling Partially Observable Cable-Driven Robot Dynamics}

\usepackage{times}
\usepackage{booktabs}
\usepackage{wrapfig}
\usepackage{graphicx}   
\usepackage{subcaption} 
\usepackage{caption}    
\usepackage[font=footnotesize]{caption}

\coltauthor{\Name{Nelson Chen}\textsuperscript{1} \Email{nelson.chen@rutgers.edu}\\
\Name{William R. Johnson} III\textsuperscript{2} \Email{will.johnson@yale.edu} \\
\Name{Rebecca Kramer-Bottiglio}\textsuperscript{2} \Email{rebecca.kramer@yale.edu} \\
\Name{Kostas Bekris}\textsuperscript{1} \Email{kostas.bekris@cs.rutgers.edu}\\
\Name{Mridul Aanjaneya}\textsuperscript{1} \Email{mridul.aanjaneya@rutgers.edu}\\
\addr \textsuperscript{1}Rutgers University, \textsuperscript{2}Yale University}

\begin{document}

\maketitle

\vspace{-.1in}
\begin{abstract}%
General-purpose simulators have accelerated the development of robots. Traditional simulators based on first principles, however, typically require full-state observability or depend on parameter search for system identification. This work presents \texttt{CableRobotGraphSim}\footnote{\url{https://nchen9191.github.io/cablerobotgraphsim/}}, a novel Graph Neural Network (GNN) model for cable-driven robots that aims to address shortcomings of prior simulation solutions. By representing cable-driven robots as graphs, with the rigid-bodies as nodes and the cables and contacts as edges, this model can quickly and accurately match the properties of other simulation models and real robots, while ingesting only partially observable inputs. Furthermore, trajectory rollout accuracy and inference speed are enhanced with prediction chunks, simultaneous multistep forward prediction. Accompanying the GNN model is a sim-and-real co-training procedure that promotes generalization and robustness to noisy real data. This model is further integrated with a Model Predictive Path Integral (MPPI) controller for closed-loop navigation, which showcases the model's speed and accuracy.%
\end{abstract}


\vspace{-0.1in}
\section{Introduction}


General-purpose simulators such as MuJoCo [\cite{mujoco}], IsaacSim [\cite{NVIDIA_Isaac_Sim}], and Drake [\cite{drake}], together with GPU-based parallelization, have made simulation central to evaluating methods and training learned controllers. Common use cases include reinforcement learning (RL) prior to real-world deployment and generating expert demonstrations for imitation learning (IL). Because robot data are scarce and expensive to collect, simulation is increasingly used to bridge this gap, especially for unconventional platforms.

\begin{wrapfigure}{r}{0.5\textwidth}
    \centering
    \vspace{-0.15in}
    \includegraphics[width=0.5\textwidth]{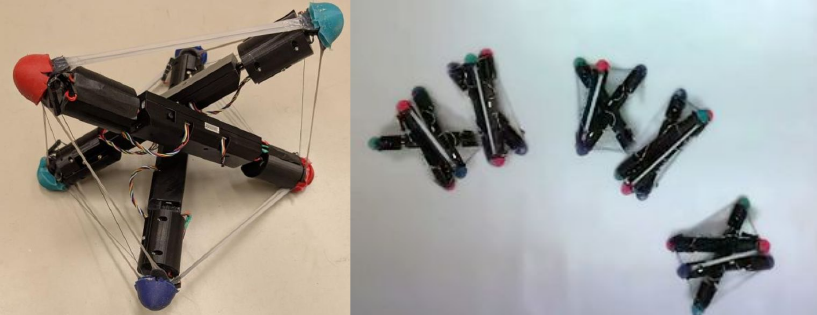}
    \vspace{-0.25in}
    \caption{Left: Static, open-source 3-bar tensegrity platform. Right: The platform rolling clockwise.}
    \vspace{-.2in}
    \label{fig:tensegrity_platform}
\end{wrapfigure}

A large category of robot platforms corresponds to cable-driven robots, which include tensegrity robots that consist of rigid rods interconnected by flexible, actuated cables. A 3-bar tensegrity used in this work and shown in Fig.~\ref{fig:tensegrity_platform}, is an instance of such structures. The properties of cable-driven robots enable applications in manipulation~[\cite{tensegrity_manipulation}], locomotion~[\cite{Sabelhaus2018DesignSA}], morphing airfoils~[\cite{airfoil}], and spacecraft landing~[\cite{Bruce2014SUPERballE}]. Their compliant and contact-rich dynamics, however, make modeling and control challenging, underscoring the need for improved simulation tools.

Traditional robot simulators have several shortcomings in this context. They face a sim-to-real gap due to oversimplified physics or incorrect system parameters, and identifying these parameters is often a manual, time-consuming process. This has motivated differentiable and learnable simulators. Differentiable analytical simulators are data-efficient but prone to local minima and restrictive assumptions, while fully learned simulators are flexible but data-hungry. Graph neural network (GNN) simulators offer a favorable middle ground, encoding structural priors via graph representations. 

Real-world data, however, is typically noisy, sparse, and only partially observable. In particular, partial observability can directly impede the use of traditional simulators that operate over a robot's full state. Soft robots are a prime example where full-state information is often not available due to their high number of degrees-of-freedom. For instance, prior work in soft robotics~[\cite{Gao_2024}] indicates that, online, there may be only access to several markers on the soft robot's surface to approximate its configuration. This is also an issue with tensegrity robots, such as the one in Fig.~\ref{fig:tensegrity_platform}, where perception methods~[\cite{tensegrity_perception}] only capture the positions of end caps and are not able to estimate the rod's twist orientation (rotation about its center-axis) or instantaneous linear and angular velocities, providing only five out of 12 dimensions of a rod's state. 

\texttt{CableRobotGraphSim}, proposed in this work, is a fully learnable graph neural network (GNN) dynamics model for cable-driven robots that is effective under partial observability, while being data-efficient. The architecture features: (i) a fully learnable GNN for simulation stability, (ii) configuration-based node features with a recurrent block to support partial observability, (iii) a cable-edge decoder for direct actuation modeling, and (iv) multi-step forward, prediction chunks to accelerate rollouts. A sim-and-real co-training procedure further enhances generalization and robustness. The method is evaluated on tensegrity robots first in simulation and then in real-world settings. Comparisons are made against existing state-of-the-art differentiable tensegrity simulators~[\cite{r2s2r,chen2024learningdifferentiabletensegritydynamics}] in a sim2sim (MuJoCo as ground truth) and real2sim setup using data from a real 3-bar tensegrity. Additionally, the model is integrated into a Model Predictive Path Integral (MPPI) controller for a simulated 6-bar tensegrity. Finally, three ablation studies assess the contributions of key architectural components, the effect of the amount of co-training data, and the impact of the size of the prediction chunk on accuracy. In summary, the contributions of this work are the following:

\vspace{0.1in}
\begin{myitem}
    \item A novel GNN architecture for learning cable-driven dynamics given partial observations.
    \item Introduction of prediction chunks for enhancing both model accuracy and inference speed for short and long horizon trajectory rollouts.
    \item A sim-and-real co-training procedure that promotes model generalization and robustness to noise.
    \item Evaluations, in simulation and reality, of the GNN applied to modeling tensegrity robots.
    \item Integration of GNN-based modeling in MPPI-based control and demonstrations of tensegrity robots navigating obstacles using this controller.
\end{myitem}

\vspace{-0.1in}
\section{Related Work}

\textbf{Data-driven, learnable simulation} methods are increasingly used in robotics to reduce the manual effort of system identification. Approaches range from differentiable physics-based simulators~[\cite{end2end_diff, robotics_diff_engine, nimblephysics}] that enable gradient computation, to purely neural network-based dynamics models~[\cite{RAISSI2019686, xu2025neuralrobotdynamics, baldan2025flowmatchingmeetspdes}]. Hybrid formulations~[\cite{Heiden2020NeuralSimAD, bianchini2023simultaneous}] combine first-principles with learned components for better generalization. A particularly effective middle ground is \textbf{graph neural network (GNN)-based dynamics modeling}~[\cite{zhang2024adaptigraphmaterialadaptivegraphbasedneural, gnn_rigid_body_sim}], which encodes structural priors through graph representations for data-efficient learning while maintaining flexibility.

For \textbf{tensegrity robots}, simulators have evolved from non-differentiable analytical ones~[\cite{tmo, stedy, motes, Paul2006DesignAC, ntrt}], which suffer from large sim-to-real gaps, to differentiable analytical simulators~[\cite{sim2sim, recurrent, r2s2r}] capable of gradient-based optimization but prone to local minima and restrictive assumptions. More recent learned simulators~[\cite{chen2024learningdifferentiabletensegritydynamics}] employ hybrid GNN approaches to model contact dynamics but require full-state observability. This work introduces a learned GNN-based simulator that addresses partial observability, while further reducing the sim-to-real gap.

In \textbf{model-based planning and control}, partial observability remains a key challenge for methods, such as model predictive control (MPC)~[\cite{Katayama04032023}], sampling-based variations~[\cite{mppi}], and iterative re-planning~[\cite{replanning, greedy-replanning}]. Prior closed-loop controllers for tensegrities~[\cite{tensegrity-mpc}] rely on simplified models or offline-generated motion primitives~[\cite{open-source-tensegrity}]. In contrast, this work demonstrates the first use of a fully learned dynamics model within an online sampling-based model predictive path integral (MPPI) controller.

Finally, insights from \textbf{robot imitation learning}~[\cite{chi2023diffusionpolicy, pi05}], notably sim-and-real co-training~[\cite{cotraining}] and action chunking~[\cite{actionchunking}], inspired part of this approach. Sim-and-real co-training aims to simultaneously use both simulation and real data to train an overall better model. Action chunking predicts a short, open-loop, temporal sequence of actions, instead of just a single action, per model run. These strategies are adapted here for dynamics learning to improve model accuracy, generalization, and simulation efficiency.

\vspace{-0.1in}
\section{Approach}

\begin{figure}[t]
    \centering
    \includegraphics[width=0.95\textwidth,height=0.4\textheight,keepaspectratio]{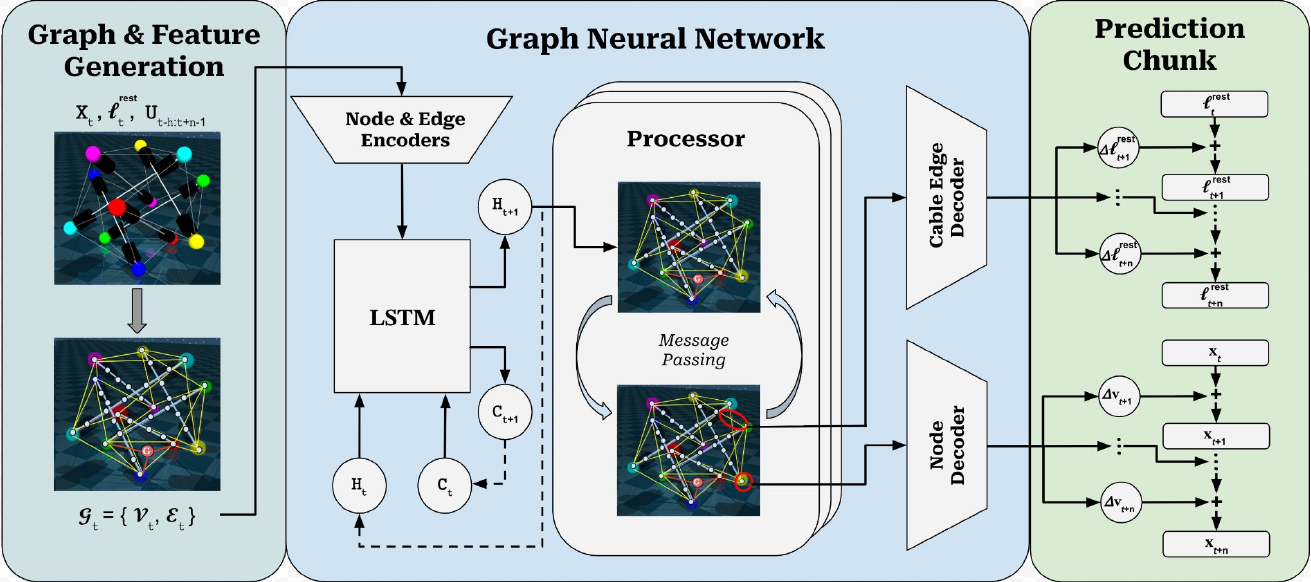}
    \vspace{-.1in}
    \caption{The three stages of a simulation step. {\bf Left}: Graph and feature generation, where state $\mathbf{X}_t$, rest length $\ell_t^{rest}$, control history $U_{t-h:t}$, and future controls $U_{t+1:t+n}$ form the graph $\mathcal{G}_t = (\mathcal{V}_t, \mathcal{E}_t)$. {\bf Middle:} $\mathcal{G}_t$ is passed through MLP encoders and an LSTM block (with hidden state $H_t$ and memory $C_t$) to produce $H_{t+1}, C_{t+1}$. $H_{t+1}$ then undergoes $L$ rounds of message passing, after which the latent node and cable-edge vectors are decoded into $n$ velocity and rest-length changes. {\bf Right:} The predicted chunk is integrated iteratively to advance $\ell_t^{rest}$ to $\ell_{t+n}^{rest}$ and $\mathbf{X}_t$ to $\mathbf{X}_{t+n}$.
        }
    \label{fig:pipeline}
    \vspace{-.2in}
\end{figure}

In cable-driven tensegrity robots, there are $K$ rigid-bodies that are connected by a set of $M$ cables, forming the robot's system topology. The state of the system is composed of the set of inner rigid-body states, where the $k$-th rigid-body at time $t$ has state $\mathbf{X}^k_t=(\mathbf{P}^k_t, \mathbf{R}^k_t, \mathbf{V}^k_t, \mathbf{\Omega}^k_t)$. Here, $\mathbf{P}^k_t$ is the position, $\mathbf{R}^k_t$ is the orientation, $\mathbf{V}^k_t$ is the linear velocity, and $\mathbf{\Omega}^k_t$ is the angular velocity. Additionally, there is a hidden state, the rest lengths of the actuated cables, $\ell^{rest}_t$, which changes as the attached motors are actuated. 
In the proposed simulator, shown in Fig. \ref{fig:pipeline}, the first step is mapping state $\mathbf{X}_t$ and controls $\mathbf{U}_t$ to the system's graph $\mathcal{G}_t$, made up of nodes $\mathcal{V}_t$ and edges $\mathcal{E}_t$. 

\vspace{-0.3in}
\begin{align}
    \mathcal{G}_t=(\mathcal{V}_t,\mathcal{E}_t)\leftarrow F(\mathbf{X}_t, \mathbf{U}_t)
\end{align}
\vspace{-0.3in}

\noindent Where $F(\cdot)$ is the graph and feature generating function. Each node's state $\mathbf{x}_t$ at time $t$ consists of their position $\mathbf{p}_t$ and linear velocity $\mathbf{v}_t$. The edges corresponding to the cables have a rest length state, $\ell^{rest}_t$.
\medskip
\noindent In the second step, the GNN, parameterized by weights $\theta$, is used to predict the changes in node velocity $\Delta \mathbf{v}_{t+1:t+n}$ and cable rest lengths $\Delta \ell^{rest}_{t+1:t+n}$, for $n$ steps forward.

\vspace{-0.3in}
\begin{align}
    (\Delta \mathbf{v}_{t+1:t+n},\;\Delta \ell^{rest}_{t+1:t+n})=\mathbf{GNN}_{\theta}(\mathcal{G}_t)
\end{align}
\vspace{-0.3in}

\noindent The third and last step is to numerically integrate the changes in velocity and rest lengths to compute the final state. In this work, the semi-explicit Euler integration scheme is chosen.

\vspace{-0.3in}
\begin{align}
    (\mathbf{v}_{t+n},\; \ell^{rest}_{t+n}) &= (\mathbf{v}_t,\;\ell^{rest}_t) + \textstyle\sum_{i=t+1}^{t+n}(\Delta \mathbf{v}_i,\;\Delta \ell^{rest}_i) \\
    \mathbf{p}_{t+n} &= \mathbf{p}_t + \Delta t \textstyle\sum_{i=t+1}^{t+n} \mathbf{v}_i
\end{align}
\vspace{-0.4in}

\subsection{GNN Model}
\vspace{-0.05in}

\textbf{Graph \& feature generation:} The graph is made up of body nodes $\mathcal{V}$, body edges $\mathcal{E}^{body}$, cable edges $\mathcal{E}^{cable}$, and contact edges $\mathcal{E}^{con}$. Each rigid body is decomposed into smaller primitive bodies such as spheres, cylinders, etc., and each is represented as a body node in the graph. Neighboring nodes that are part of the same rigid body are connected by a body edge. Here, the ground is represented by a single body node. Together, this forms $\mathcal{V}=\{V_i\}$ and $\mathcal{E}^{body}=\{E^{body}_{ij}\}$. Next, each cable is represented by edges, forming $\mathcal{E}^{cable}=\{E^{cable}_{ij}\}$. Contact edges are dynamically assigned to connect body nodes and the ground node if the distance between the two are below a user-defined threshold, forming $\mathcal{E}^{con}=\{E^{con}_{ij}\}$. 


Two feature sets distinguish this model from prior GNN-based simulators. First, configuration-based features include (i) relative distance to a designated node in the robot graph, (ii) relative distance to the rigid body’s CoM, and (iii) the body frame's $z$-axis. While many learned simulators~[\cite{gns-rigid}] only use velocity inputs to preserve translational equivariance, configuration-based features aid inference under partial observability by providing implicit state context. Second, for cable edges, a sequence of past $h$ and future $n$ control signals ($U_{t-h:t+n}$) are included to support both actuation learning and multi-step prediction.

\smallskip
\noindent \textbf{GNN architecture:} In prior work [\cite{chen2024learningdifferentiabletensegritydynamics}], the cable and motor dynamics are computed analytically and combined with a GNN to learn contact dynamics, forming a hybrid approach. This work instead opts to be fully learnable. The authors have observed, in their experiments, that the hybrid approach becomes unstable when the inputs are only partial observations. This is likely due to the cables being a stiff linear system, where the cables create a negative feedback loop with the GNN. The stiff cables would greatly compound the error of the GNN, and in turn, would rapidly take the state out of the GNN's training distribution.

Following other established GNN architectures~[\cite{gns-rigid, gnn_rigid_body_sim}], this work also employs the encode-process-decode structure. In the \textbf{GNN encoder}, there is similarly multi-layered perceptrons (MLP), per node and edge type, to encode the raw features to a larger latent space. In this work, a long-short term memory (LSTM) block is added to the encoder to retain temporal hidden states $H^0_{i}$ and memory information $C_{i}$. The usage of hidden states and memory encodes history information and is a common method to deal with partial observability.

\vspace{-0.3in}
\begin{align}
    (H^0_{i},C_{i})&=\mathbf{LSTM}(\;\mathbf{MLP}^{enc}_{node}(V_{i}), H_i,C_{i})\\
    \mathrm{E}^{\epsilon,0}_{ij} &= \mathbf{MLP}_{\epsilon}^{enc}(E^{\epsilon}_{ij}) \;\;\; \epsilon\in\{body,cable,contact\}
\end{align}
\vspace{-0.3in}


\noindent In the \textbf{GNN processor}, $L$ message passes and node updates are executed. For the $l^{th}$ message pass, a set of MLPs corresponding to each edge type, and an MLP for node updates, are used. First, messages are computed per edge based on the nodes it connects and its current latent vector. Second, these messages are aggregated via a sum operation across same edge types, and concatenated with other aggregated edge messages, per node. Last, the node's latent vector and the aggregated edges are concatenated, and passed to an MLP to update the node's updated latent vector:

\vspace{-0.25in}
\begin{align}
    \mathrm{E}^{\epsilon,l}_{ij}&=\mathbf{MLP}^{MP}_l\left(H^{l-1}_i, H^{l-1}_j, \mathrm{E}^{\epsilon,l-1}_{ij}\right) \;\;\; \epsilon\in\{body, cable, contact\}\\
    H^l_i&=\mathbf{MLP}^{update}_l\left(H^{l-1}_i, \sum{\mathrm{E}^{body,l}_{ij}}, \sum{\mathrm{E}^{cable,l}_{ij}}, \sum{\mathrm{E}^{con,l}_{ij}}\right)
\end{align}
\vspace{-0.25in}

\noindent After the processor, two \textbf{GNN decoders}, each an MLP, are used to take the latest node and cable edge latent vectors and map them to $n$-step velocity changes $\Delta \mathbf{v}_{t+1:t+n}$ and cable rest-length changes $\Delta \ell^{rest}_{t+1:t+n}$. The cable-edge decoder acts as a learnable cable actuation model by taking the latest cable-edge latent vector as input. This latent vector contains information about the properties of the cable, relevant control signals, and node effects on the cable motor.

\vspace{-0.27in}
\begin{align}
    \Delta \mathbf{v}_{t+1:t+n} = \mathbf{MLP}^{dec}_{node}\left(H^L_i\right)\;\;\;\;\;\; \Delta \ell^{rest}_{t+1:t+n}=\mathbf{MLP}^{dec}_{cable}\left(\mathrm{E}^{cable,L}_{ij}\right)
\end{align}
\vspace{-0.25in}


\noindent \textbf{Prediction chunk:} Traditional simulators use a timestep $\Delta t$ to advance the state from $t$ to $t+1$, requiring $T/\Delta t$ steps to simulate a rollout of duration $T$. Increasing $\Delta t$ reduces the number of steps but degrades numerical accuracy. In contrast, the proposed method predicts $n$ future steps from state $t+1$ to $t+n$ simultaneously, henceforth referred to as a prediction chunk. This approach (i) preserves fine timestep size discretization while achieving an $n$-fold speedup, (ii) provides dense supervision through intermediate ground truths, and (iii) enables efficient long-range propagation via a more compact computational graph. The decoders output $\Delta \mathbf{v}_{t+1:t+n}$ and $\Delta \ell^{rest}_{t+1:t+n}$, from which the final states are iteratively integrated using the chosen scheme.

\medskip
\noindent \textbf{Loss function:} To train the GNN, as well as the internal cable actuation decoder, a single standard mean-squared-error (MSE) loss $\mathcal{L}$, over the predicted and ground-truth (GT) node velocity changes and rest length changes, averaged over the prediction chunk size $n$, is used: 

\vspace{-0.25in}
\begin{align}
    \mathcal{L} &= \frac{1}{n}\sum_{\tau=1}^{n}\left[\frac{\rho_n}{B_{n}} \sum_{i\in \mathcal{V}}{\left(\Delta \mathbf{v}_{i,\tau}^{pred} - \Delta \mathbf{v}_{i,\tau}^{GT}\right)^2} + \frac{\rho_c}{B_{c}} \sum_{j \in \mathcal{E}^{cable}}{\left(\Delta \ell_{j,\tau}^{rest,pred} - \Delta \ell_{j,\tau}^{rest,GT}\right)^2}\right]
\end{align}
\vspace{-0.2in}

\noindent where $B_{n}$ and $B_{c}$ are the number of nodes and actuated cables in a training batch, respectively, and $\rho_n$ and $\rho_c$ are user-defined weights balancing the two.

\vspace{-0.1in}
\subsection{Sim and real data co-training}
\vspace{-0.05in}

For real robots, their data are often noisy and sparse. This is especially true for unconventional robots that do not have large communities or resources to collect abundant amounts of data. However, neural networks are very data hungry, causing learned models to be prone to overfitting and to be sensitive to noise. This is why differentiable analytical simulators are the go-to choice due to the physics principles embedded in the model. A line of research that recently emerged in robot learning is the idea of co-training~[\cite{cotraining}] control policies with a mix of simulation and real data. For this work, sim-and-real co-training has been adapted for learning dynamics. The inclusion of simulation data provides the GNN with a foundational source of physics knowledge, making it robust to noise in the real data. 


\begin{wrapfigure}{l}{0.55\textwidth}
    \centering
    \vspace{-.2in}
    \includegraphics[width=0.55\textwidth]{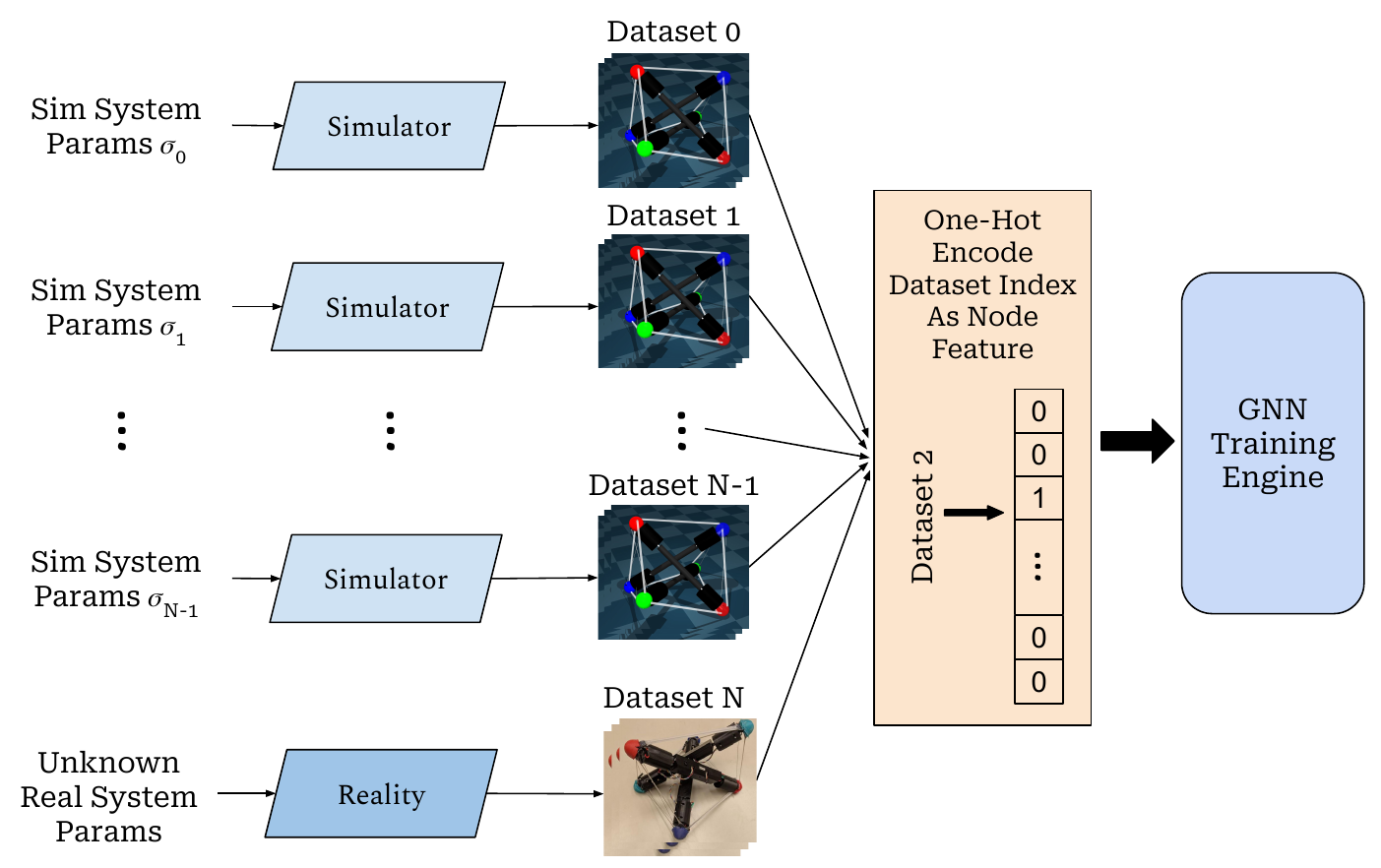}
    \vspace{-0.25in}
    \caption{Sim-and-real co-training. Multiple simulation datasets are generated by varying system parameters, and are pooled with a real dataset of unknown system parameters. Data points are distinguished by a one-hot encoding node feature of the dataset enumerated index.}
    \vspace{-.15in}
    \label{fig:cotraining}
\end{wrapfigure}

\noindent To generate the simulation co-training data, select a simulator or simulation platform that can roughly model the system of interest, but cannot easily perform system identification. Next, identify a set of relevant system parameters to vary and generate multiple parameter sets either by grid or random sampling. Then, take the controls and initial states of the real robot trajectories, and run rollouts in simulation for each set of system parameters $\sigma_i$ to generate dataset $\mathcal{D}_i$. 

Because the real data's system parameters are not actually known, the simulation datasets' system parameters cannot be used by the GNN. Consequently, a naive mixing of all data sources would result in the GNN becoming confused, i.e., the same input state results in different outputs. To combat this, the simulation datasets are enumerated from $0$ to $N-1$, and the real dataset is enumerated as the $N^{th}$ dataset. Finally, this information is passed to the GNN as a one-hot encoded node feature. This process can be seen in Fig. \ref{fig:cotraining}.

\vspace{-0.1in}
\subsection{MPPI integration}
\vspace{-0.05in}

To demonstrate the effectiveness of the proposed GNN model, this paper integrates the full GNN model into an MPPI controller. Furthermore, the MPPI controller also acts as a diverse data generator that is used to further train the GNN model, which subsequently, improves the controller. This can be repeated for multiple iterations to finetune both the model and controller to a particular task.

\medskip
\noindent \textbf{Background} MPPI is a sampling-based MPC method for stochastic optimal control. Given a state-transition model $f_{sim}$, MPPI samples $K$ noisy input control sequences $\{U^k\}$. Next, $K$ trajectories $\{\mathcal{T}^k\}$ are generated by simulating $\{U^k\}$ and initial state $X_0$ with $f_{sim}$ over a short time horizon $T$.

\vspace{-0.25in}
\begin{align}
    \mathcal{T}^k=\left[\mathbf{X}_0,f_{sim}\left(\mathbf{X}_0,u^k_0\right),...,f_{sim}\left(\mathbf{X}_{T-1},u^k_{T-1}\right)\right]
\end{align}
\vspace{-0.3in}

\noindent With a cost function $\mathcal{C}$ to be minimized, importance sampling weights $w^k$ can be computed with an inverse exponential:

\vspace{-0.4in}
\begin{align}
    w^k=\frac{1}{\eta}\exp\left(-\frac{1}{\beta}(\mathcal{C}\left(\mathcal{T}^k\right)-\mu)\right), \;\;\; \sum{w^k}=1
\end{align}
\vspace{-0.24in}

\noindent where $\eta$ is a normalization factor, $\beta$ is the inverse temperature parameter, and $\mu=\min_kC^k$. Finally, the optimal control sequence $U^*$ is approximated with the weighted average

\vspace{-0.3in}
\begin{align}
    U^*=\sum{w^k V^k}
\end{align}
\vspace{-0.3in}

\noindent The first control of $U^*$ is then executed and the MPPI process is repeated again to compute the next optimal control sequence. The GNN model and MPPI controller naturally work well together, since the GNN can natively simulate all rollouts in parallel on GPUs. In this work, $f_{sim}$ is substituted with the proposed trained GNN model. The cost function used is

\vspace{-0.3in}
\begin{align}
    \mathcal{C}(\mathbf{X}_t)= \alpha_1d_{L_1}(\mathbf{X}_t)+\alpha_2/d_{collision}(\mathbf{X}_t)
\end{align}
\vspace{-0.3in}

\begin{figure}[t]
  \centering
  \includegraphics[width=0.97\textwidth]{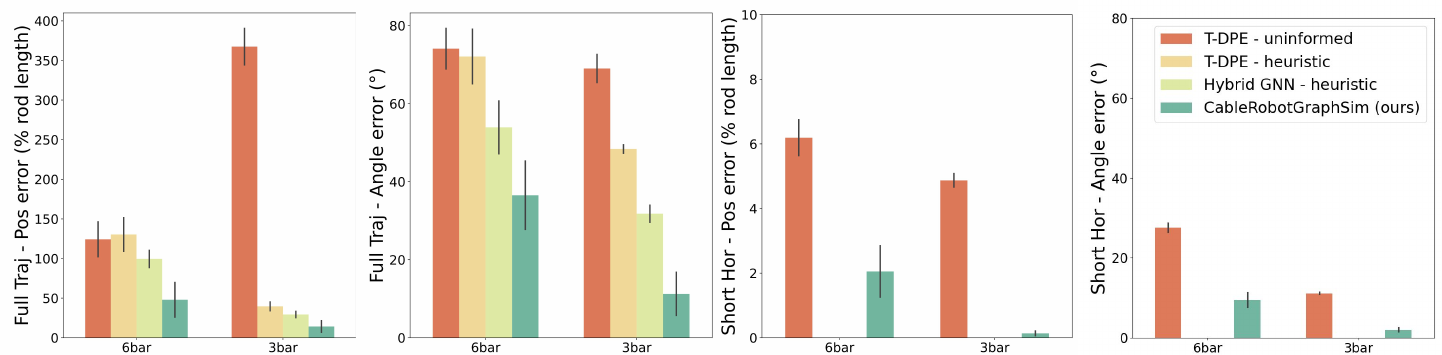}
  \vspace{-0.1in}
  \caption{Sim2sim evaluation for full trajectories and short horizons, as well as a 3-bar and 6-bar tensegrity.}
  \label{fig:s2s}
  \vspace{-0.2in}
\end{figure}

\noindent where $d_{L_1}(\cdot)$ is the \textit{obstacle-aware} 2D manhattan distance of the current center of mass (CoM) to the goal, $d_{collision}(\cdot)$ is the minimum distance between the robot end caps and the obstacles, and $\alpha_1$ and $\alpha_2$ are the respective cost weights.

The MPPI controller serves as an effective data generator due to its stochastic sampling and ability to produce meaningful locomotion. This is advantageous for learned models, which require broad coverage of the state–control space to avoid overfitting and bias. Previously, tensegrity control relied on hand-crafted gaits~[\cite{johnson-tensegrity}] or random actions that failed to induce motion. In this work, the MPPI controller operates in a loop: (i) train a GNN model, (ii) execute the controller with the updated model over a task, and (iii) record and add resulting trajectories to the dataset. This process is repeated across iterations while tracking model accuracy and task-level metrics such as success rate and completion time.

\vspace{-0.1in}
\section{Experimental Results}

For the sim2sim and real2sim experiments, the proposed approach is compared against \cite{chen2024learningdifferentiabletensegritydynamics}, a hybrid analytical and GNN tensegrity simulator, and \cite{r2s2r}, a tensegrity differentiable physics engine, henceforth referred as Hybrid GNN and T-DPE, respectively. Each baseline adapts two cable attachment variants: a heuristic scheme and an uninformed scheme. The heuristic attachment, developed in prior work \cite{r2s2r}, specifies how to attach cables to the surface of the rod end caps when the platform is at rest. The uninformed attachment places cables at the center of the end caps, simplifying the model by removing the need for the missing rod twist orientations. Because the heuristic attachment is only valid at rest, its corresponding baselines cannot be used in an online, closed-loop setting, which requires the simulator to run from any valid non-rest state. All methods are trained and tested on a single machine with an Nvidia RTX 4090 GPU and an AMD Ryzen 7950 16-core 4.5 GHz CPU, and all simulators were built and executed using PyTorch and PyGeometric.

\medskip
\noindent \textbf{Evaluation metrics:} For the following evaluation results, there are several common metrics used. These metrics are CoM MSE $\mathbf{e}_{pos}$ (normalized by rod length $L_{rod}$), and  angular error $\mathbf{e}_{rot}$ over the rods' center-axis $\hat{\mathbf{r}}$, both averaged over the trajectory length $T$. Formally, the error metrics are

\vspace{-0.25in}
\begin{align}
    \mathbf{e}_{pos}=\frac{1}{L_{rod} T}\sum{\left(\mathbf{P}_i^{GT}-\mathbf{P}_i^{pred}\right)^2},\;\; \textbf{e}_{rot}=\frac{1}{T}\sum{\cos^{-1}{\left(\mathbf{\hat{r}}_i^{GT}\cdot \mathbf{\hat{r}}_i^{pred}\right)}}
\end{align}
\vspace{-0.2in}

\noindent These metrics are applied in two different settings, full trajectory $(\mathbf{e}^{full}_{pos}, \mathbf{e}^{full}_{rot})$ and short horizon $(\mathbf{e}^{SH}_{pos}, \mathbf{e}^{SH}_{rot})$. A full trajectory evaluation compares a full simulation rollout against the ground truth. The short horizon evaluation samples $N_{SH}$ random initial states from a trajectory, rolls the samples out for a short horizon $T_{SH}$, and compares the simulation rollouts against the ground-truth rollouts. The short horizon metrics give insight into how the model would perform in receding horizon controllers like MPPI.

\smallskip
\noindent {\bf Simulation-to-Simulation:} In the sim2sim experiments, there are six different simulation datasets generated with MuJoCo as the ground-truth environment. Each is generated with a different set of system parameters, specifically the coefficient of friction and the contact solver's stiffness parameter. Each dataset has nine trajectories ranging from 30 to 60 seconds each, with a train-test split of six to three trajectories, respectively. This was done both for a 3-bar tensegrity and a 6-bar tensegrity. The 3-bar used human-engineered gait patterns, while the 6-bar used a mix of passive throws, random controls, and MPPI-based trajectories. The proposed method and the baselines are trained and tested on each dataset, and the mean and standard deviation of the errors are shown in Fig.~\ref{fig:s2s}. As a reminder, the heuristic variants can only be used from the initial rest states of each trajectory, so they cannot be applied in the short horizon setting. Moreover, the uninformed Hybrid GNN baseline resulted in unstable behavior and diverged for all metrics, so numerical evaluation is omitted. Fig.~\ref{fig:s2s} shows that the proposed method yielded strong performance gains over the baselines.

\begin{table}[h]
\centering
\small
\vspace{-.1in}
\begin{tabular}{ccccc}
\toprule
 & Full Traj & Full Traj & Short Horizon & Short Horizon\\  
Model& Pos Error (\%) & Rot Error (\textdegree) & Pos Error (\%) & Rot Error (\textdegree) \\
\midrule
Hybrid GNN - uninformed & Unstable & Unstable & Unstable & Unstable \\
Hybrid GNN - heuristic & 91.12 & 35.29 & N/A & N/A \\
T-DPE - uninformed & 465.59 & 60.32 & 9.59 & 28.76\\
T-DPE - heuristic & 102.76 & 38.48 & N/A & N/A  \\
CableRobotGraphSim (ours) & \textbf{48.95} & \textbf{26.53} & \textbf{3.95} & \textbf{7.031}\\
\bottomrule
\end{tabular}
\vspace{-0.1in}
\caption{Real2sim evaluation results on a real 3-bar tensegrity platform.}
\label{tab:r2s}
\vspace{-0.1in}
\end{table}

\noindent {\bf Real-to-Simulation:} In the real2sim experiments, the goal is to best match the simulator to reality. A dataset of nine trajectories is generated by running the human-engineered gait patterns with a real 3-bar tensegrity. The data are recorded using a tensegrity perception algorithm~[\cite{tensegrity_perception}], which only captures the rods' end cap positions, missing the rods' twist orientation and instantaneous velocities. As shown in Table \ref{tab:r2s}, the proposed method shows strong relative performance compared to the alternatives. The real2sim errors are overall larger than the sim2sim errors. This is to be expected, since simulation dynamics are a simplified model of reality. Another source of error is that the ground truth used here is state estimation from the perception algorithm that may be noisy, which would also be reflected in the real2sim results.

\begin{figure}[b]
  \vspace{-.35in}
  \includegraphics[width=0.97\textwidth]{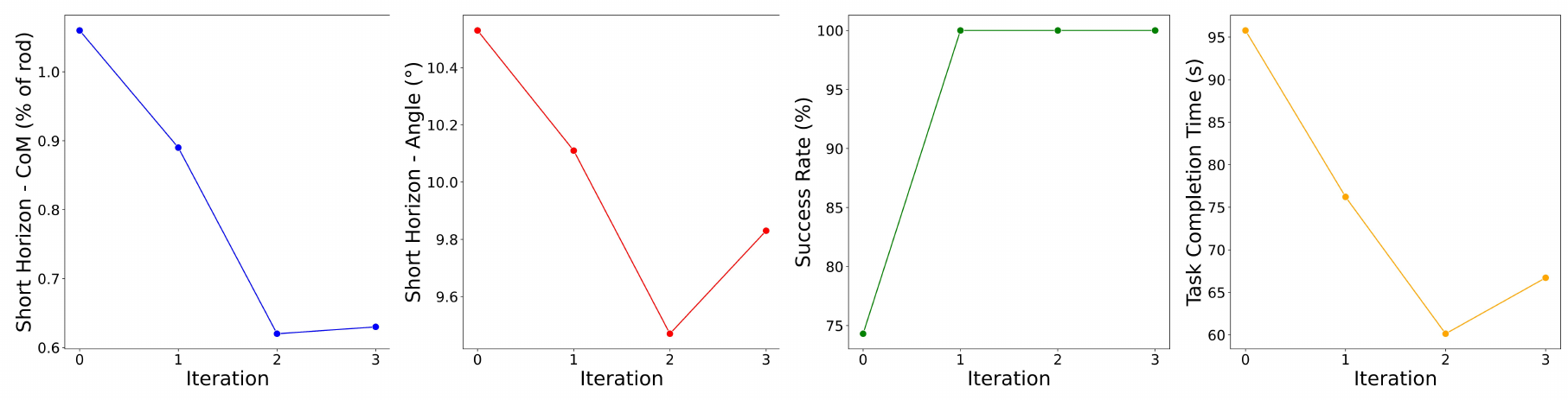}
  \vspace{-.05in}
  \caption{Running MPPI controller in a loop of model training, task execution, and new data generation for model retraining over three iterations. Positional and rotational errors are evaluated on a pooled test set from all iterations. Success rate and task completion time are measured per iteration.}
  \label{fig:sim_mppi}
\end{figure}


\smallskip
\begin{wrapfigure}{r}{0.45\textwidth}
    \centering
    \vspace{0.05in}
    \includegraphics[width=0.45\textwidth]{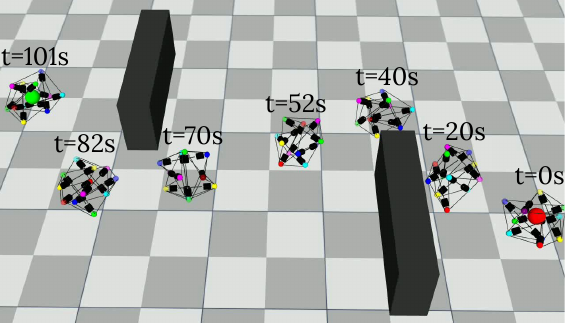}
    \vspace{-0.3in}
    \caption{S-shaped obstacle course for navigating from the red to the green sphere.}
    \vspace{-.15in}
    \label{fig:mppi_obs_course}
\end{wrapfigure}
\begin{wrapfigure}{l}{0.33\textwidth}
    \centering
    \vspace{-0.15in}
    \includegraphics[width=0.33\textwidth]{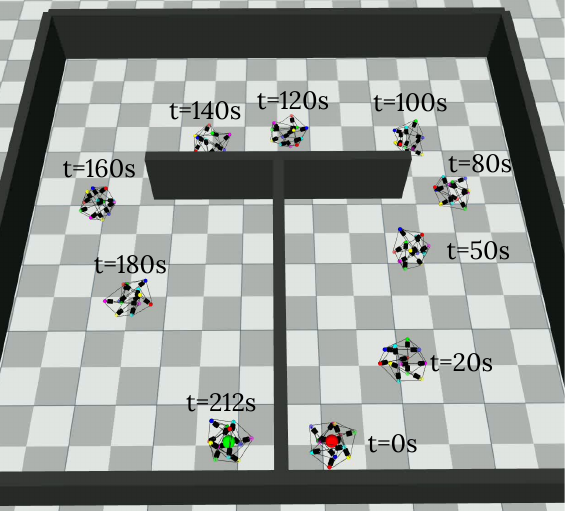}
    \vspace{-0.3in}
    \caption{Maze map for navigating from the red to the green sphere.}
    \vspace{-.2in}
    \label{fig:mppi_maze}
\end{wrapfigure}

\noindent {\bf MPPI and controller-in-the-loop data collection:} The GNN is integrated with an MPPI controller as an internal state-transition model. The controller is also used for active data collection. Experiments are conducted on a simulated 6-bar tensegrity navigating an obstacle course (Fig.~\ref{fig:mppi_obs_course}). An initial model (iteration~0) is trained using trajectories from random controls and passive drops, followed by three data collection and retraining iterations. For each iteration, success rate (goal reached within 120~s) and task completion time are measured. After the final iteration, all test data are aggregated into a single evaluation set, and each iteration's model is assessed for short-horizon, positional and rotational error (Fig.~\ref{fig:sim_mppi}). Both the model and controller improve over iterations, with slight regression at iteration~3, likely due to model capacity. The final model, when deployed with the MPPI in a previously unseen maze (Fig.~\ref{fig:mppi_maze}) can complete the navigation task successfully.

\smallskip
\noindent {\bf Ablations:} Three ablations assess the contributions of individual components. The first (Table~\ref{tab:model_comp_ablation}) incrementally transitions the Hybrid GNN baseline to the proposed architecture by: (i) converting it to a fully learned GNN, (ii) adding configuration-based features, (iii) incorporating a recurrent block, and (iv) introducing an actuation model (analytical or learned). Experiments (i)-(iii) use ground-truth actuation outputs, while (iv) compares analytical versus learned actuation given control inputs. On sim2sim data, the full proposed setup achieves the best performance without requiring ground-truth actuation.

\begin{table}[t]
\centering
\small
\vspace{0.2in}
\begin{tabular}{ccccc}
\toprule
 & Full Traj & Full Traj & Short Horizon & Short Horizon\\ 
Model& Pos Error (\%) & Rot Error (\textdegree) & Pos Error (\%) & Rot Error (\textdegree) \\
\midrule
(baseline) Hybrid GNN & Unstable & Unstable & Unstable & Unstable \\
(i) Fully Learned GNN (F-GNN) & $20.6 \pm 26.14$  & $47.1 \pm 27.03$  & $0.11 \pm 0.04$  & $4.91 \pm 0.99$ \\
(ii) F-GNN + Config Feats (FC-GNN) & $1.15 \pm 1.13$&	$15.43 \pm 15.54$	&$0.1 \pm 0.04$	&$4.6 \pm 1.21$\\
(iii) FC-GNN + Recurrent (FCR-GNN) & $\mathbf{1.04 \pm 1.3}$&	$\mathbf{7.61 \pm 4.72}$&	$\mathbf{0.06 \pm 0.02}$&	$\mathbf{3.24 \pm 0.88}$ \\
\midrule
FCR-GNN + analytical motor &$4.49 \pm 2.21$	&$31.18 \pm 7.22$	&$0.1 \pm 0.04$	&$5.14 \pm 0.66$ \\
(iv) FCR-GNN + learned motor (ours) &$\mathbf{1.1 \pm 0.9}$&	$\mathbf{14.02 \pm 7.84}$&	$\mathbf{0.06 \pm 0.02}$	&$\mathbf{3.11 \pm 0.48}$  \\
\bottomrule
\end{tabular}
\vspace{-0.1in}
\caption{GNN architecture component ablation study. The first four rows compare GNN components using ground-truth motor outputs. The last two rows compare an analytical vs. a learned motor model.}
\vspace{-0.1in}
\label{tab:model_comp_ablation}
\end{table}

\begin{figure}[h]
  \vspace{-0.05in}
  \centering
  \includegraphics[width=0.97\textwidth]{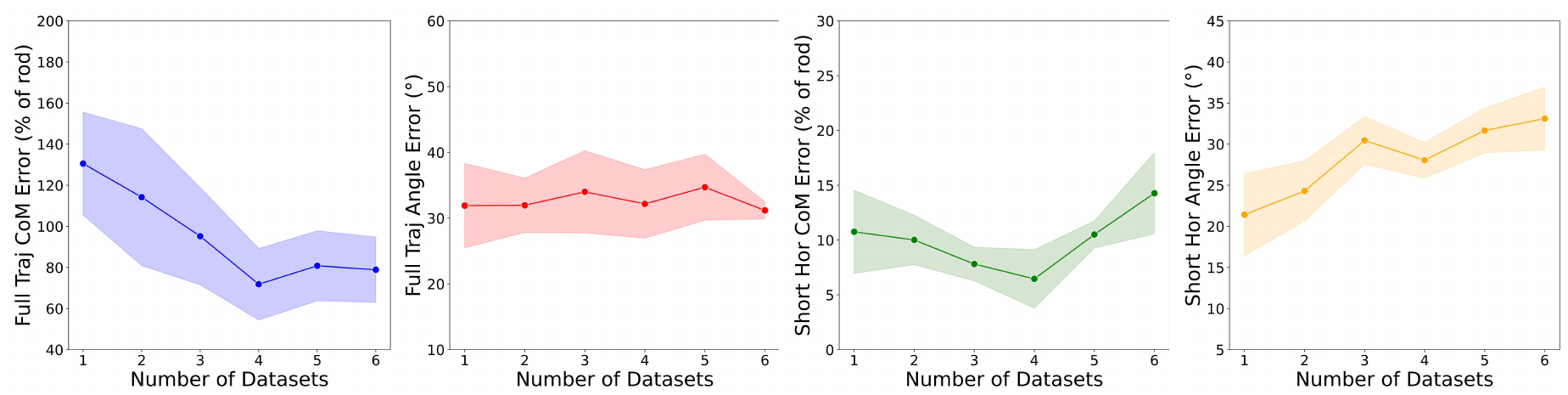}
  \vspace{-0.1in}
  \caption{Sim-and-real co-training ablation study for varying number of simulation datasets.}
  \label{fig:co_training}
  \vspace{-0.1in}
\end{figure}

\noindent The second ablation (Fig.~\ref{fig:co_training}) varies the number of simulation datasets used for sim-and-real co-training, evaluating each resulting model on real test data. Models trained without simulation data were unstable, likely due to the sparsity and noise of real data, and are omitted. Positional errors improve with up to four simulation datasets and then degrade, suggesting that simulation data acts as a regularizer that becomes too strong in excess. Rotational errors plateau or slightly increase, likely because the position-based loss weighs positional accuracy more heavily, as the angular component is bounded by the robot's geometry.

\begin{figure}[h]
  \vspace{-0.1in}
  \includegraphics[width=\textwidth]{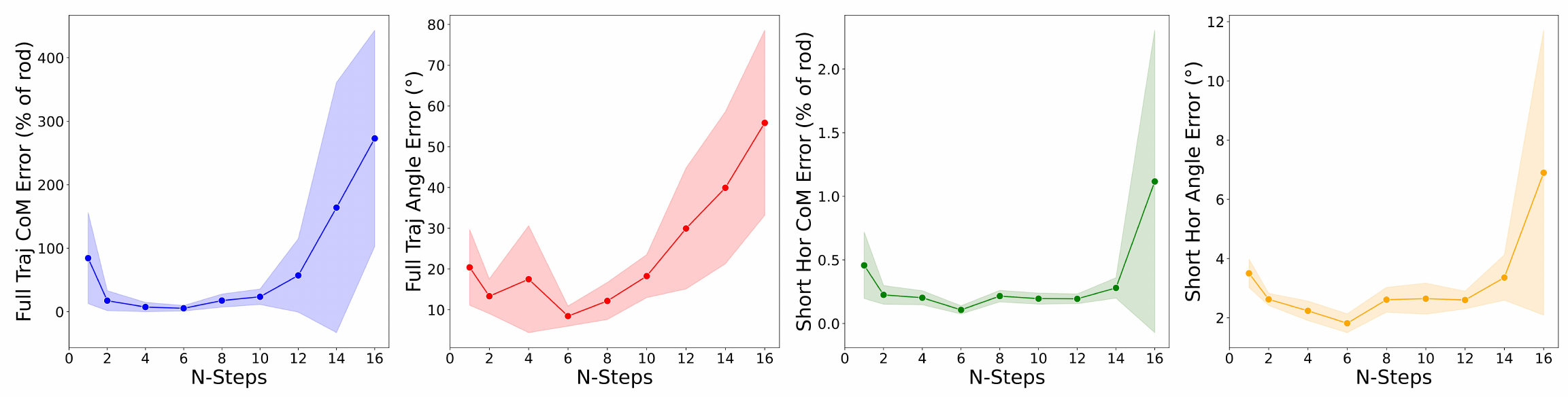}
  \vspace{-.3in}
  \caption{Prediction chunk size ablation over the number of forward time steps per simulation call.}
  \vspace{-.3in}
  \label{fig:multi_pred_steps}
\end{figure}

\noindent The third ablation examines the effect of the number of forward prediction steps, varying from one to 16 in increments of two. As shown in Fig.~\ref{fig:multi_pred_steps}, performance peaks at six steps. This is unexpected, since shorter horizons are typically easier to predict. A likely explanation is that learned models suffer from compounding errors as each auto-regressive step moves the state further out of distribution, so increasing steps per prediction reduces total inferences per rollout. Beyond six steps, accuracy degrades again. While model size is not explored here, evidence from visual-language-action (VLA) models suggests larger models could support longer horizons, hinting at an optimal pairing of model size and prediction length for rollout efficiency.

\vspace{-0.05in}
\section{Discussion}
\vspace{-0.05in}
\texttt{CableRobotGraphSim} demonstrates the flexibility of learned simulators, while addressing challenges related to noise and data sparsity. By combining a graphical representation, a novel GNN architecture for cable-driven robots, and a sim-and-real co-training scheme, the model effectively handles partial observability as well as limited and lower quality data. Experiments on cable-driven tensegrity platforms show strong performance against baselines in both sim2sim and real2sim settings. The model is further integrated with an MPPI controller, which allows the successful completion of navigation tasks while generating additional training data. 

At the same time, the current model assumes flat terrain. Future work will extend it to uneven surfaces for more complex navigation.  A natural next step is to apply similar GNN-based frameworks to other robotic systems that can benefit from a learned graphical representation. This will also allow a deeper understanding of the benefits of sim-and-real co-training and multi-step prediction chunking for robot modeling purposes.

\acks{Nelson Chen and Mridul Aanjaneya were supported in part by the National Science Foundation (NSF) under awards CCF-2110861, IIS-2132972, IIS-2238955, and CCF-2312220 as well as a research gift from Red Hat, Inc. William R. Johnson III and Rebecca Kramer-Bottiglio were supported by the NSF under award IIS-195522.  Kostas Bekris was supported in part by NSF under award IIS-1956027.}

\bibliography{l4dc2026-sample}

\end{document}